\documentclass{article}

\usepackage[american]{babel}
\PassOptionsToPackage{numbers, compress}{natbib}

\usepackage[final]{nips_2016}

\usepackage[utf8]{inputenc} 
\usepackage[T1]{fontenc}    
\usepackage[breaklinks=true,colorlinks,bookmarks=false]{hyperref}       
\usepackage{booktabs}       
\usepackage{amsfonts}       
\usepackage{microtype}      
\usepackage{graphicx}
\usepackage{multirow}
\usepackage{subfig}

\graphicspath{{figure/}}

\title{Mask-CNN: Localizing Parts and Selecting Descriptors for Fine-Grained Image Recognition}

\author{
  Xiu-Shen Wei \qquad Chen-Wei Xie \qquad Jianxin Wu\\
  National Key Laboratory for Novel Software Technology, Nanjing University \\
  163 Xianlin Avenue, Qixia District, Nanjing 210023, China \\
  \texttt{\{weixs,xiecw,wujx\}@lambda.nju.edu.cn}
}

\begin{document}

\maketitle

\begin{abstract}
Fine-grained image recognition is a challenging computer vision problem, due to the small inter-class variations caused by highly similar subordinate categories, and the large intra-class variations in poses, scales and rotations. In this paper, we propose a novel end-to-end Mask-CNN model without the fully connected layers for fine-grained recognition. Based on the part annotations of fine-grained images, the proposed model consists of a fully convolutional network to both locate the discriminative parts (e.g., head and torso), and more importantly generate object/part masks for selecting useful and meaningful convolutional descriptors. After that, a four-stream Mask-CNN model is built for aggregating the selected object- and part-level descriptors simultaneously. The proposed Mask-CNN model has the smallest number of parameters, lowest feature dimensionality and highest recognition accuracy when compared with state-of-the-arts fine-grained approaches. 
\end{abstract}

\section{Introduction} \label{sec:intro}

Fine-grained recognition tasks such as identifying the species of a bird, have been popular in computer vision. Since the categories are all similar to each other, different categories can only be distinguished by slight and subtle differences, which makes fine-grained recognition a challenging problem. Compared to the general object recognition tasks, fine-grained recognition benefits more from learning critical parts of the objects, which helps discriminate different subclasses and align objects of the same class~\cite{Azizpour12ECCV, Shao16CVPR, Di15CVPR, Ning14ECCV, Yu16TIP}.

In the deep learning era, a straightforward way to represent parts is to use the deep convolutional features/descriptors. The convolutional descriptors contain more localized (i.e., parts) information compared to the feature of the fully connected layers (i.e., whole image). In addition, these deep descriptors are known to correspond to mid-level information, e.g., object parts~\cite{Zeiler13ICCV}. All the previous part-based fine-grained approaches, e.g.,~\cite{Shao16CVPR, Di15CVPR, Ning14ECCV, Yu16TIP}, directly used the deep convolutional descriptors and encoded them into a single representation, without evaluating the usefulness of the obtained object/part deep descriptors. By using powerful convolutional neural networks~\cite{cnn12}, we may not need to select useful dimensions inside feature vectors, as what we do to hand-crafted features~\cite{Angela12NIPS, Wu14CVPR}. However, since most deep descriptors are not useful or meaningful for fine-grained recognition, it is necessary to select useful deep convolutional descriptors. Recently, selecting deep descriptors sheds its light on the fine-grained image retrieval task~\cite{Wei16}. Moreover, it is also beneficial to fine-grained image recognition.

\begin{figure}[t]
 \centering
 \subfloat[Part annotations]  { \includegraphics[width=0.32\columnwidth]{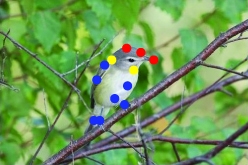} \label{fig:maskGT_2} }
 \qquad
 \subfloat[Part polygons] { \includegraphics[width=0.32\columnwidth]{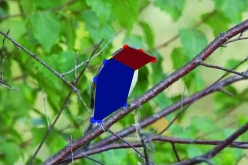} \label{fig:maskGT_3} }
 \caption{We generate the convex polygons (in~\ref{fig:maskGT_3}) for the bird's head and torso based on the part annotations (red, blue and yellow dots in~\ref{fig:maskGT_2}). Other pixels are treated as background. The two yellow part key points (i.e., nape and throat) are included in both head torso. (Best if viewed in color.)} \label{fig:maskGT}
\end{figure}

\begin{figure}[t]
 \centering
 {\includegraphics[width=0.98\columnwidth]{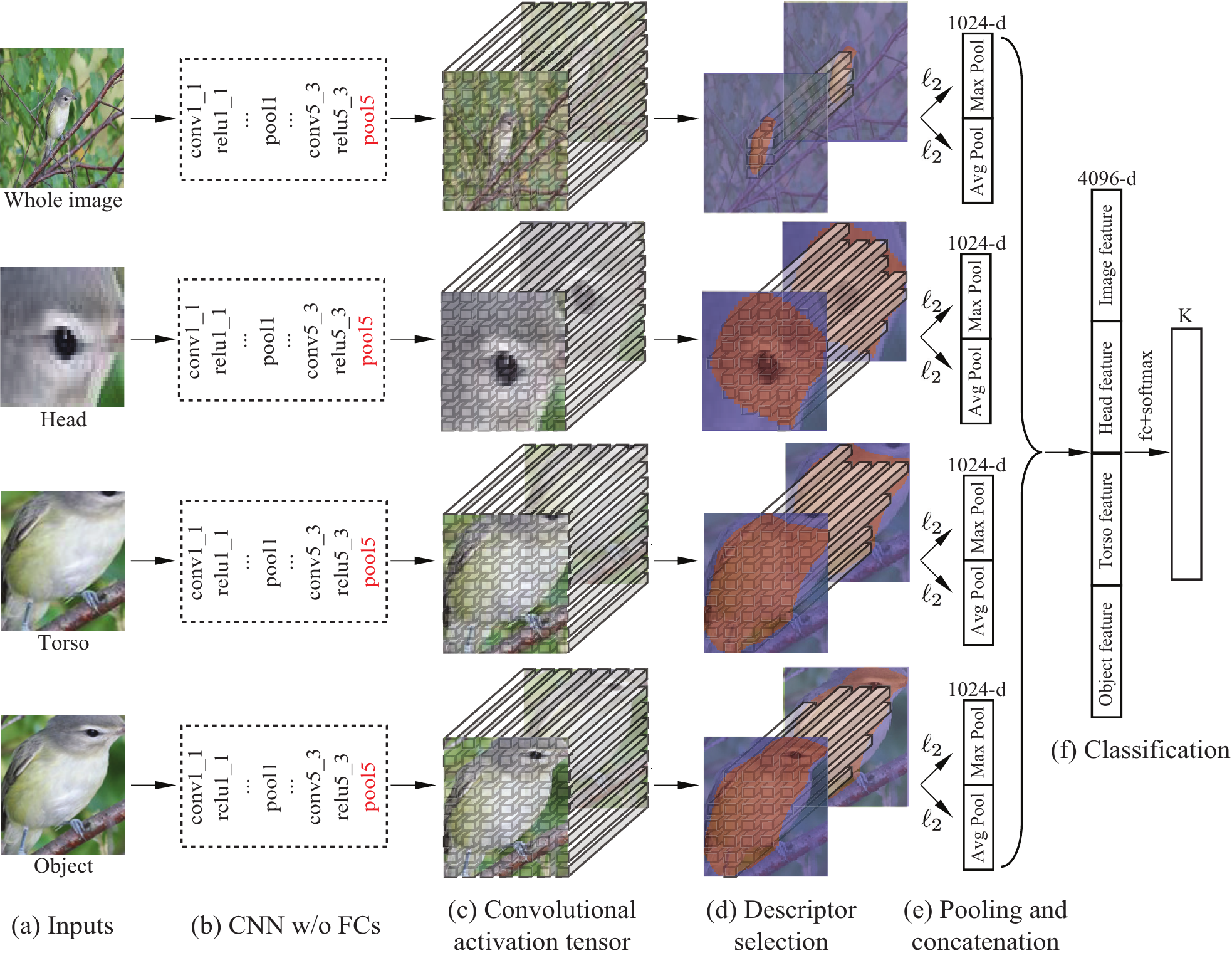}}
 \caption{Architecture of the proposed four-stream Mask-CNN. The four streams correspond to the whole image, head, torso and object images/patches, respectively. Note that we removed the fully connected layers. As illustrated in this figure, thanks to the descriptor selection scheme, a large number of descriptors corresponding to background can be discarded by M-CNN, which is beneficial to fine-grained recognition. (This figure is best viewed in color.)}  \label{fig:framework}
\end{figure}

In this paper, by developing a novel deep part detection and descriptor selection scheme, we propose an end-to-end Mask-CNN (M-CNN) model which discards the fully connected layers for fine-grained recognition. We only require the part annotations and image-level labels during the training time. In M-CNN, given the part annotations, we firstly separate them into two point sets. One set corresponds to the head part of the fine-grained bird image, and the other is for the torso. Then, the smallest convex polygons that cover each point set are returned as the ground-truth mask, as shown in Fig.~\ref{fig:maskGT}. The other pixels are background. By treating part localization as a three-class segmentation task, we leverage fully convolutional networks (FCN)~\cite{Jonathan15CVPR} to generate masks in the testing time for both localizing parts and selecting useful deep descriptors, which does not use any annotation during testing. After getting these two part masks, we combine them to form the object. Based on these object/part masks, a four-stream Mask-CNN (image, head, torso, object) is built for joint training and aggregating the object-level and part-level cues simultaneously. The architecture of the proposed four-stream M-CNN is shown in Fig.~\ref{fig:framework}. In each stream of M-CNN, we discard the fully connected layers of CNNs. In the last convolutional layer, an input image is represented by multiple deep descriptors. In order to select useful descriptors to keep only those corresponding to the object, the pre-learned object/part masks by FCN are used. After that, the selected descriptors of each stream are both averaged and max pooled into 512-d feature vectors. The standard $\ell_2$-normalization is followed. Finally, the feature vectors of these four streams are concatenated, and then a classification (fc+softmax) layer is added for end-to-end joint training.

We validate the proposed four-stream M-CNN on the popular Caltech-UCSD Birds-200-2011~\cite{WahCUB200_2011} dataset, in which we achieved 85.5\% classification accuracy. We also get accurate part localization (84.62\% for head and 89.83\% for torso). The key advantages and major contributions of the proposed M-CNN model are:
\begin{description}
 \item[1.] To the best of our knowledge, Mask-CNN is the first end-to-end model that selects deep convolutional descriptors for object recognition, especially for fine-grained image recognition.
 \item[2.] We present a novel and efficient part-based four-stream model for fine-grained recognition. We discard the fully connected layers, and the proposed M-CNN is computationally and storage efficient. Comparing with state-of-the-art methods, M-CNN has the least parameters and smallest feature dimensionality (60.49M and 8,192-d), respectively. At the same time, it achieves 85.4\% classification accuracy on CUB200-2011, which is the highest among existing methods. With the SVD whitening method, our feature representation can be compressed to 4,096-d, and meanwhile improve the accuracy to 85.5\%.
 \item[3.] The part localization performance of the proposed model outperforms other part-based fine-grained approaches which requires additional bounding boxes. In particular, M-CNN is about 10\% higher than state-of-the-art for head localization.
\end{description}

\section{Related Work} \label{sec:related}

Fine-grained recognition is a challenging problem and has recently emerged as a hot topic. During the past few years, a number of effective fine-grained recognition methods have been developed in the literature~\cite{Shao16CVPR, Max15NIPS, Di15CVPR, Tsung-Yu15ICCV, Ning14ECCV, Yu16TIP}. We can roughly categorize these methods into three groups. The first group, e.g.,~\cite{Max15NIPS, Tsung-Yu15ICCV}, attempted to learn a more discriminative feature representation by developing powerful deep models for classifying fine-grained images. The second group aligned the objects in fine-grained images to eliminate pose variations and the influence of camera position, e.g.,~\cite{Steve14BMVC, Gavves14IJCV, Di15CVPR}. The last group focused on part-based representations, because it is widely acknowledged that the subtle difference between fine-grained images mostly resides in the unique properties of object parts.

For the part-based fine-grained recognition methods, \cite{Azizpour12ECCV, Di15CVPR, Ning14ECCV} used both bounding boxes of the birds and part annotations during training to learn an accurate part localization model. Then, based on these detected parts, different CNNs are fine-tuned using the detected parts separately. To ensure satisfactory localization results, they even used bounding boxes in the testing phase. In contrast, our method only need part annotations for training, and do not need any supervision during testing. Moreover, our four-stream M-CNN is a unified framework for capturing object- and part-level information simultaneously. Some other part-based methods considered a weakly supervised setting, in which they categorize fine-grained images with only image-level labels, e.g.,~\cite{Marcel15ICCV, Tianjun15CVPR, Xiaopeng16CVPR, Yu16TIP}. As will be shown by our experiments, classification accuracy of M-CNN is significantly higher than these weakly supervised methods. Meanwhile, the model size of M-CNN is the smallest among all state-of-the-art methods, which make it efficient to train.

Besides, there are also fine-grained recognition methods based on segmentation, e.g.,~\cite{Shao16CVPR, Jonathon15CVPR}. The most significant difference between them and M-CNN is: these methods only use segmentation to localize the whole object~\cite{Jonathon15CVPR} or parts~\cite{Shao16CVPR}, while we further select useful deep convolutional descriptors using the masks from segmentation. Among them, the part-stacked CNN model~\cite{Shao16CVPR} is the most related work to ours. In~\cite{Shao16CVPR}, part-stacked CNN requires both bounding box and part annotations in training, and even needed the bounding boxes during testing. Within the image patch cropped using the bounding box, \cite{Shao16CVPR} treated the image crop around each of the fifteen part key points as 15 segmentation foreground classes, and used FCN to solve the 16-classes segmentation task. After obtaining the trained FCN, it localized these part point positions in the last convolutional layer. Then, deep activations corresponding  to the fifteen parts and the whole object were stacked together. Fully connected layers were used for classification. Comparing with part-stacked CNN, M-CNN only needs to localize two main parts (head and torso), which makes the segmentation problem much easier and more accurate. M-CNN achieves high localization accuracy, as will be shown in Table~\ref{table:IOU}. Meanwhile, as demonstrated in~\cite{Shao16CVPR}, using all the fifteen part activations cannot lead to better classification accuracy. Besides, M-CNN's accuracy on CUB200-2011 is 1.8\% higher than that of~\cite{Shao16CVPR} using the same baseline network, although we use less annotations in training and do not use any annotation in testing (cf. Sec.~\ref{sec:sota}).

\section{The Mask-CNN Model} \label{sec:method}

In this section, we present the proposed four-stream Mask-CNN (M-CNN) model. Firstly, we adopt a fully convolutional network (FCN)~\cite{Jonathan15CVPR} to generate the object/part masks for locating object/parts, and more importantly selecting deep descriptors. Then, based on these masks, the four-stream M-CNN is built for joint training and capturing both object- and part-level information.

\subsection{Learning Object and Part Masks}\label{sec:learnmask}

The fully convolutional network (FCN)~\cite{Jonathan15CVPR} is designed for pixel-wise labeling. FCN can take an input image with any resolution and produce an output of corresponding dimensions. In our method, we use FCN to not only localize the object and parts in fine-grained images, but also treat the segmentation predictions as the object and parts masks for the later descriptor selection process.

\begin{figure}[t]
 \centering
 \includegraphics[width=\columnwidth]{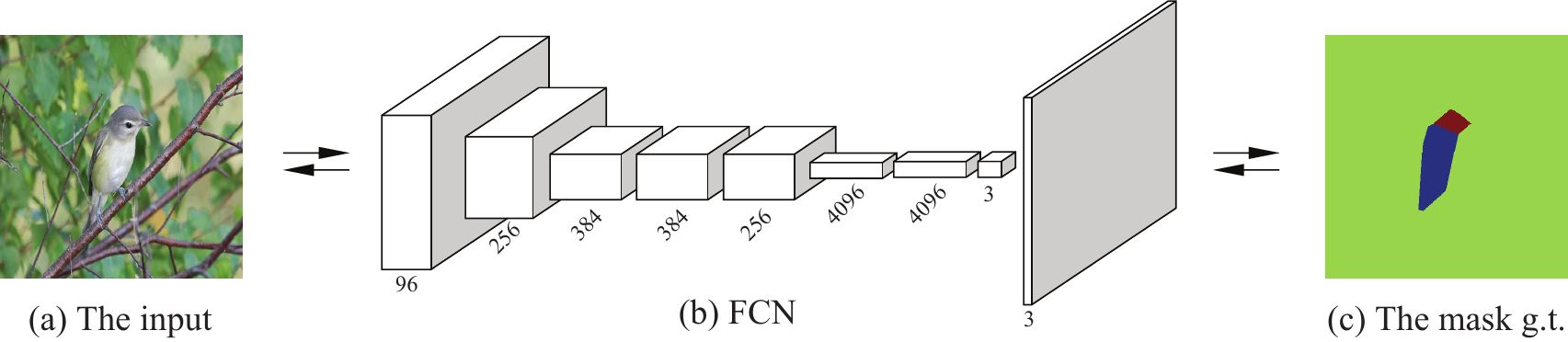}
 \caption{Demonstration of the mask learning procedure by FCN~\cite{Jonathan15CVPR}. (Best viewed in color.)}  \label{fig:fcn}
\end{figure}

Each fine-grained image in the CUB200-2011~\cite{WahCUB200_2011} dataset is annotated with part annotations, i.e., fifteen part key points. As shown in Fig.~\ref{fig:maskGT}, we split these key points into two sets, including the head key points (i.e., the beak, forehead, crown, left eye, right eye, nape and throat) and torso key points (i.e., the back, breast, belly, left leg, right leg, left wing, nape, right wing, tail and throat). Based on the key points, two ground-truth of part masks are generated. One is the {\em head mask}, which corresponds to the smallest convex polygon covering all the head key points. The other is the {\em torso mask}, which is the smallest convex polygon covering the torso key points. In Fig.~\ref{fig:maskGT}, the red polygon is the head mask, and the blue one is for torso. The rest of the image is background. Therefore, we model the part mask learning procedure as a three-class segmentation problem. For effective training, all the training and testing fine-grained images are with their original resolutions. Then, we crop a $384\times 384$ image patch in the middle of the original image as the inputs. The mask learning network architecture is shown in Fig.~\ref{fig:fcn}. In our experiments, we adopted FCN-8s~\cite{Jonathan15CVPR} for learning and predicting part masks.

\begin{figure}[t]
\centering
 {\includegraphics[width=\columnwidth]{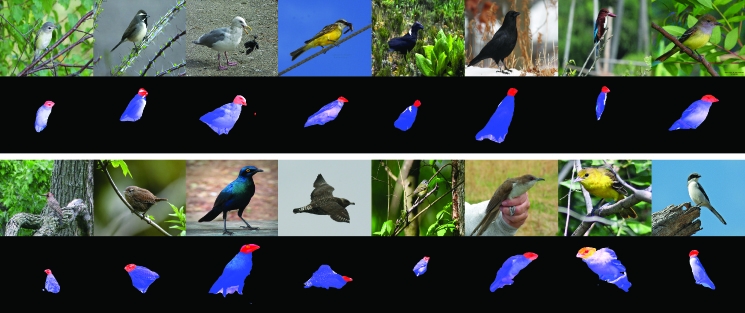}}
 \caption{Sixteen random samples of predicted part masks from the testing set. In these figures, we overlay the part mask predicted by FCN (the head highlighted in red and the torso in blue) onto the original images. The pixels predicted as background are in black. (Best viewed in color.)}  \label{fig:predmask}
\end{figure}

During the FCN inference, without using any annotation, three class heat maps (in the same size as the original input image) are returned for each image. We randomly choose some qualitative examples of the predicted part masks, and show them in Fig.~\ref{fig:predmask}. In these figures, the learned masks are overlaid onto the original images. The head part is highlighted in red, and the torso is in blue. The predicted background pixels are in black. As can be seen from these figures, even though the ground-truth part masks are not very accurate, the learned FCN model is able to return more accurate part masks. Meanwhile, these part masks can also localize the part positions. Quantitative results of part localization and object segmentation will be reported in Sec.~\ref{sec:experi_partacc} and Sec.~\ref{sec:experi_seg}, respectively. 

Both part masks, if accurately predicted, will benefit the later deep descriptor selection process and the final fine-grained classification. Therefore, during both training and testing, we will use the predicted masks for both part localization and descriptor selection in M-CNN. We also combine the two masks to form a mask for the whole object, which is called the {\em object mask}.

\subsection{Training Mask-CNN}\label{sec:jointlearn}

After obtaining the object and part masks, we build the four-stream M-CNN for joint training. The overall architecture of the proposed model is presented in Fig.~\ref{fig:framework}. We take the whole image stream as an example to illustrate the pipeline of each stream in M-CNN. 

The inputs of the whole image stream are the original images resized with $h\times h$. In our experiments, we report the results for $h=224$ and $h=448$, respectively. The input images are fed into a traditional convolutional neural network, but the fully connect layers are discarded. That is to say, the CNN model used in our proposed M-CNN only contains convolutional, ReLU and pooling layers, which greatly brings down the M-CNN model size. Specifically, we use VGG-16~\cite{vgg16} as the baseline model, and the layers before $\rm pool_5$ are kept (including $\rm pool_5$). We obtain a $7\times 7\times 512$ activation tensor in $\rm pool_5$ if the input image is $224\times 224$. Therefore, we have 49 deep convolutional descriptors of 512-d, which also correspond to $7 \times 7$ spatial positions in the input images. Then, the learned object mask (cf. Sec.~\ref{sec:learnmask}) is firstly resized to $7\times 7$ by the nearest interpolation, and then used for selecting useful and meaningful deep descriptors. As illustrated in Fig.~\ref{fig:framework} (c) and (d), the descriptor should be kept when it locates in the object region. If it locates in the background region, that descriptor will be discarded. In our implementation, the mask is set as a binary matrix, in which 1 stands for keeping and 0 is for discarding. We implement the selection process as an element-wise product operation between the convolutional activation tensor and the mask matrix, which is similar to the element-sum summarize operation in FCN~\cite{Jonathan15CVPR}. Therefore, the descriptors located in the object region will remain, while the other descriptors will become zero vectors.

For these selected descriptors, in the end-to-end M-CNN learning process, we both average and max pool them into two 512-d feature vectors, respectively. Then, the $\ell_2$-normalization is followed for each of them. After that, we concatenate them into an 1024-d feature as the final representation of the whole image stream.

The streams for head and torso have similar processing steps as the whole image one. However, different from the inputs of the whole image stream, we generate the input images of the head and torso streams as follows. After obtaining the two part masks (i.e., the head and torso masks), we use the part masks as the part detectors to localize the head part and torso part in the input images. For each part, we return the smallest rectangle bounding box which contains the part mask regions. Based on the rectangle bounding box, we crop the image patch which acts as the inputs of the part stream. The two streams in the middle of Fig.~\ref{fig:framework} show the head and torso streams in M-CNN. The last stream is the object stream, which crops the image patch by combining the two part masks into an object mask. Thus, its inputs are the main object (i.e., bird) detected by our FCN segmentation network. The inputs of these three streams are all resized into $224\times 224$ in our experiments.

In the classification step shown in Fig.~\ref{fig:framework} (f), the final 4,096-d image representation is the concatenation of the whole image, the head, the torso and the object features. The last layer of M-CNN is a 200-way classification (fc+softmax) layer for classification on the CUB200-2011 dataset. The four stream M-CNN is learned end-to-end, with the parameters of four CNNs learned simultaneously. During training M-CNN, the parameters of the learned FCN segmentation network are fixed.

\section{Experiments} \label{sec:experiments}

In this section, we firstly describe the experimental settings and implementation details. Then, we report the classification accuracy and present discussions about the proposed M-CNN model. Finally, the performance of part localization and object segmentation will also be provided.

\subsection{Dataset and Implementation Details}

The empirical evaluation is performed on the widely-used fine-grained benchmark Caltech-UCSD 2011 bird dataset~\cite{WahCUB200_2011}. This dataset contains 200 bird categories, and each category has roughly 30 training images. We follow the training and testing splitting included with the dataset. In the training phase, the fifteen part annotations are adopted for generating the part masks' ground-truth, and meanwhile the image-level labels are used for the end-to-end M-CNN joint training. We need no supervision signals (e.g., part annotations or bounding boxes) when testing. 

The proposed Mask-CNN model and FCN used for generating masks are implemented using the open-source library MatConvNet~\cite{matconvnet}. In our experiments, after getting the learned part masks, we firstly generate the image patches of birds' head, torso and object as described in Sec.~\ref{sec:jointlearn}. Then, to facilitate the convergence of four stream CNNs, each single stream corresponding to the whole image, head, torso and object is fine-tuned on its input images separately. The CNNs used in each stream is initialized by the popular VGG-16 model~\cite{vgg16} pre-trained on ImageNet. In addition, we double the training data by horizontal flipping for all the four streams. After fine-tuning on each stream, as shown in Fig.~\ref{fig:framework}, the joint training of four-stream M-CNN is performed. Dropout is not used in M-CNN. At the test time, we average the predictions of the image and its flipped copy, and output the class with the highest score as the prediction for a test image. In addition, directly using the softmax predictions results is a slight drop in accuracy compared to logistic regression (LR), which is consistent with the observations in~\cite{Tsung-Yu15ICCV}. Therefore, in the following, the reported results of M-CNN are all achieved by one-vs-all logistic regression~\cite{REF08a} on the extracted features (4096-d) with the default hyper-parameter $C_{\rm LR} = 1$.

\subsection{Classification Accuracy and Comparisons}

We report the classification accuracy on the CUB200-2011 dataset of the proposed four-stream M-CNN model, and compare with the baseline methods and state-of-the-art methods in the literature.

\subsubsection{Baseline Methods}

In order to validate the effectiveness of the descriptor selection process in M-CNN, we perform two baseline methods which are also based on the proposed four-stream architecture. Different from our M-CNN, these two baseline methods do not contain the descriptor selection part, i.e., the processing shown in Fig.~\ref{fig:framework} (d).

The first baseline method employ the traditional fully connected layers to conduct classification for each stream, which is called ``4-stream FCs''. In ``4-stream FCs'', we replace the (b) to (e) parts of each stream in Fig.~\ref{fig:framework} with a CNN containing fully connected layers (i.e., VGG-16 with only {\rm fc8} removed). Thus, the generated feature in the last layer of each stream is a 4,096-d single vector. The rest procedure is also to concatenate the four 4,096-d features into the final one with 16,384-d, and to learn a 200-way classification (fc+softmax) layer on the 16,384-d image representation.

The second baseline is similar to the proposed M-CNN. The most prominent difference is that it discards the descriptor selection part, i.e., the processing in Fig.~\ref{fig:framework} (d). Thus, the convolutional deep descriptors of $\rm pool_5$ in each stream are directly average and max pooled, and then $\ell_2$-normalized, respectively. Therefore, we call it the ``4-stream Pooling''. The remaining procedures are the same as the proposed M-CNN.

Table~\ref{table:baseline} presents the comparison of classification accuracy on the CUB200-2011 dataset, where the input images of the whole image stream are $224\times 224$. The proposed M-CNN achieves the best classification accuracy rate. Due to the missing of descriptor selection, ``4-stream Pooling'' is about 1\% lower than M-CNN. The ``4-stream FCs'' baseline method has the lowest accuracy. Its lower accuracy might be caused by the fully connected layers, which may have caused overfitting.

\begin{table}[t]
 \caption{Comparison with the baseline methods on CUB200-2011.} \label{table:baseline}
 \small
 \centering
 \begin{tabular}{c|c|c}
  \hline
  4-stream FCs & 4-stream Pooling & The proposed 4-stream M-CNN \\
  \hline
  81.1\% & 82.2\% & \textbf{83.1\%} \\
  \hline
 \end{tabular}
\end{table}

\subsubsection{Comparisons with state-of-the-art methods} \label{sec:sota}
 
The classification accuracy of the proposed four-stream M-CNN and state-of-the-art methods on CUB200-2011 are presented in Table~\ref{table:acc}. For fair comparison, we only report the results when they do not use part annotations in testing. 

As aforementioned, when all the inputs are of size $224\times 224$, the accuracy of the proposed four-stream M-CNN model is 83.1\%. Following~\cite{Tsung-Yu15ICCV}, we change the input images of the whole image stream to $448\times 448$ pixels, which improves the classification performance by 2.1\%. We also resize the input images of the object stream to $448\times 448$. But the accuracy is slightly lower than before.

Moreover, as the ensemble of multiple layers can boost the final performance~\cite{Jonathan15CVPR, Wei16}, after joint training, we extract the deep descriptors from the $\rm relu_{5\_2}$ layer which is three layers in front of $\rm pool_5$. Then, the predicted part masks are also used to select the corresponding descriptors of the four streams. Similar to the pooling and concatenation processes done for $\rm pool_5$, we can obtain another 4,096-d image representation of $\rm relu_{5\_2}$. After that, we combine it with the $\rm pool_5$ one into a 8,192-d feature vector (called ``4-stream M-CNN$^+$'' in Table~\ref{table:acc}), which achieves the best classification accuracy 85.4\% on CUB200-2011. Additionally, we compress the 8,192-d feature vector to 4,096 by SVD whitening. It can reduce the dimensionality, and meanwhile improve the accuracy to 85.5\%.

Specifically, because part-stacked CNN~\cite{Shao16CVPR} used the Alex-Net model~\cite{cnn12}, we also build another four-stream M-CNN based on Alex-Net. The accuracy of our four-stream M-CNN (Alex-Net) is 78.0\%. It is 1.8\% higher than that of~\cite{Shao16CVPR}. Moreover, in Alex-Net based four-stream M-CNN, the number of parameters is only 9.74M, and the final feature vector is only 2,048-dimensional .

\begin{table}[t]
 \caption{Comparison of classification accuracy on CUB200-2011 with state-of-the-arts methods.} \label{table:acc}
 \centering
 \setlength{\tabcolsep}{2pt}
 \small
 \begin{tabular}{*{6}{c|}r|r|c}
  \hline
  \multirow{2}{*}{Method} & \multicolumn{2}{c|}{Train phase} & \multicolumn{2}{c|}{Test phase} & \multirow{2}{*}{Model} & \multirow{2}{*}{$\sharp$ para.~~~} & \multirow{2}{*}{Dim.~~} & \multirow{2}{*}{Acc.} \\
  \cline{2-5} & BBox & Parts & BBox & Parts & & & \\
  \hline
  Part-Stacked CNN~\cite{Shao16CVPR} & $\checkmark$ & $\checkmark$ & $\checkmark$ & & Part-Stacked CNN$\times 1$ & 130.80M & \textbf{4,096} & 76.2\% \\
  PB R-CNN with BBox~\cite{Ning14ECCV} & $\checkmark$ & $\checkmark$ & $\checkmark$ & & Alex-Net$\times 3$  & 173.03M & 12,288 & 76.4\% \\
  Deep LAC~\cite{Di15CVPR} & $\checkmark$ & $\checkmark$ & $\checkmark$ & & Alex-Net$\times 3$ & 173.03M & 12,288 & 80.3\% \\
  PB R-CNN~\cite{Ning14ECCV} & $\checkmark$ & $\checkmark$ & & & Alex-Net$\times 3$ & 173.03M & 12,288 & 73.9\% \\  
  Pose Normalized CNNs~\cite{Steve14BMVC} & $\checkmark$ & $\checkmark$ & & & Alex-Net$\times 3$ & 173.03M & 13,512 & 75.7\% \\ 
  Co-Segmentation~\cite{Jonathon15CVPR} & $\checkmark$ & & & &VGG-19$\times 2$ & 287.30M & 126,976  & 82.0\%   \\ 
  Two-Level~\cite{Tianjun15CVPR} & & & & &VGG-16$\times 1$ & 138.35M & 16,384  & 77.9\%   \\
  Weakly supervised FG~\cite{Yu16TIP} & & & & &VGG-16$\times 1$ & 138.35M & 262,144  & 79.3\%   \\
  Constellations~\cite{Marcel15ICCV} & & & & &VGG-19$\times 1$ & 143.65M & 208,896 & 81.0\%  \\
  Bilinear~\cite{Tsung-Yu15ICCV} & & & & &VGG-16 and VGG-M & 73.67M & 262,144 & 84.1\% \\
  Spatial Transformer CNN~\cite{Max15NIPS} & & & & & ST-CNN (inception)$\times 4$ & 62.68M & \textbf{4,096} & 84.1\% \\
  PDFS~\cite{Xiaopeng16CVPR} & & & & & VGG-16$\times 1$ & 138.35M & 69,632 & 84.5\% \\
  \hline
  \hline
  {Our 4-stream M-CNN (224)} & & $\checkmark$& & & {VGG-16 (w.o. FCs)$\times 4$}& \textbf{59.67M} & \textbf{4,096} & {83.1\%} \\
  {Our 4-stream M-CNN (448)} & & $\checkmark$& & & {VGG-16 (w.o. FCs)$\times 4$}& \textbf{59.67M} & \textbf{4,096} & {85.2\%} \\
  {Our 4-stream M-CNN$^+$ (448)} & & $\checkmark$ & & & {VGG-16 (w.o. FCs)$\times 4$}& 60.49M & {8,192} & \textbf{85.4\%} \\
  \hline
 \end{tabular}
\end{table}

\subsection{Part Localization Results}\label{sec:experi_partacc}
Except for the qualitative part localization results shown in Sec.~\ref{sec:learnmask}, in this section, we quantitatively assess the localization correctness using the Percentage of Correctly Localized Parts (PCP) metric. As reported in Table~\ref{table:IOU}, the metrics are the percentage of parts (i.e., the head and torso) that are correctly localized with a $>$50\% IOU with the ground-truth part bounding boxes as generated in~\cite{Di15CVPR, Ning14ECCV}. By comparing the results of PCP for torso, our method outperforms part-based R-CNN~\cite{Ning14ECCV} and strong DPM~\cite{Azizpour12ECCV} by a large margin. However, because we do not use any supervision in testing, the localization performance is lower than the one of Deep LAC~\cite{Di15CVPR} which used the bounding boxes during testing. In addition, for the head localization task which is more challenging than the torso one, even though our method just uses part annotations in training, the head localization performance (84.62\%) is still significantly higher than the other methods. 

\begin{table}[t]
 \caption{Comparison of part localization performance on the CUB200-2011 dataset.} \label{table:IOU}
 \small
 \centering
 \begin{tabular}{c|c|c|c|c|c|c}
  \hline
  \multirow{2}{*}{Method} & \multicolumn{2}{c|}{Train phase} & \multicolumn{2}{c|}{Test phase} & \multirow{2}{*}{Head} & \multirow{2}{*}{Torso} \\
  \cline{2-5} & BBox & Parts & BBox & Parts & & \\
  \hline
  Strong DPM~\cite{Azizpour12ECCV} & $\checkmark$ & $\checkmark$ & $\checkmark$ & & 43.49\% & 75.15\% \\
  Part-based R-CNN with BBox~\cite{Ning14ECCV} & $\checkmark$ & $\checkmark$ & $\checkmark$ & & 68.19\% & 79.82\% \\
  Deep LAC~\cite{Di15CVPR} & $\checkmark$ & $\checkmark$ & $\checkmark$ & & 74.00\% & \textbf{96.00\%} \\
  Part-based R-CNN~\cite{Ning14ECCV} & $\checkmark$ & $\checkmark$ & & & 61.42\% & 70.68\% \\  
  \hline
  Ours & & $\checkmark$ & & & \textbf{84.62\%} & 89.83\% \\
  \hline
 \end{tabular}
\end{table}

\subsection{Object Segmentation Performance}\label{sec:experi_seg}

\begin{figure}[t]
 \centering
 \includegraphics[width=0.9\columnwidth]{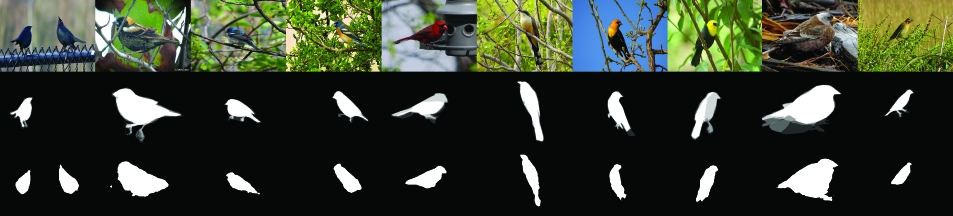}
 \caption{Examples of segmentation results. The first row is the original fine-grained images. The second row is the corresponding segmentation ground-truth. The last row is the predicted results. Note that although the segmentation ground-truth only annotates one bird, there are two birds in the first image and M-CNN correctly finds both.}  \label{fig:segm}
\end{figure}

Because the CUB200-2011 dataset also supplies the object segmentation ground-truth, we can directly test the learned object masks on the segmentation metric. Fig.~\ref{fig:segm} shows qualitative segmentation results. Our method based on FCN is generally able to segment the foreground object well, but understandably has trouble to segment the finer birds' parts, e.g., claws and beak. Since our goal is not to segment objects, we do not perform any refinement as pre-processing or post-processing. Moreover, we evaluate the segmentation performance quantitatively by the common semantic segmentation metric mean IU (pixel accuracy and region intersection over union) of the ground truth foreground object wit the predicted object masks. It is 72.41\% on the testing set. In fact, a better segmentation result will lead to better predicted object/part masks, and also benefit the final classification. To further improve the classification accuracy, some pre-processing methods, e.g., GrabCut~\cite{grabcut}, are worth trying to obtain better mask ground-truth than the convex polygons in Fig.~\ref{fig:fcn} (c).

\section{Conclusion} \label{sec:conclusion}

In this paper, we presented the benefits of selecting deep convolutional descriptor in object recognition, especially fine-grained image recognition. By developing the descriptor selection scheme, we proposed a novel end-to-end Mask-CNN (M-CNN) model without the fully connected layers to not only accurately localize object/parts, but also generate object/part masks for selecting deep convolutional descriptors. After aggregating the selected descriptors, the object-level and part-level cues were encoded by the proposed four-stream M-CNN model. Mask-CNN not only achieved 85.5\% classification accuracy on CUB200-2011, but also had the least parameters and the lowest dimensional feature representations.

In the future, we plan to solve the part detection problem of M-CNN in the weakly supervised setting, in which we only require the image-level labels. Thus, it will require far less labeling effort to achieve comparable classification accuracy. In addition, another interesting direction is to explore the benefits of descriptor selection for general object categorization.


\small
\bibliographystyle{nips2016}
\bibliography{MCNN}

\end{document}